%% file: main.tex
\documentclass[letterpaper]{article} 
\usepackage[submission]{aaai2026}    

\usepackage{times}    
\usepackage{helvet}   
\usepackage{courier}  
\usepackage[hyphens]{url}  
\usepackage{graphicx} 
\usepackage{natbib}   
\usepackage{caption}  
\usepackage{algorithm}
\usepackage{algorithmic}
\usepackage{tabularx}
\usepackage{booktabs}
\usepackage{multirow}
\usepackage{cuted}
\usepackage{comment}
\usepackage{pgfplots}
\usepackage{xcolor}
\usepackage[accsupp]{axessibility}
\usepackage{enumitem}
\usepackage{array}
\usepackage{xspace}
\usepackage{soul}
\usepackage[theorems,skins,breakable]{tcolorbox}
\usepackage{tikz}
\usepackage{amsmath}
\usepackage{newfloat}
\usepackage{listings}

\pgfplotsset{compat=1.18}

\newcommand{\eg}{e.g.\xspace}

\definecolor{mypurple}{RGB}{100,42,150}
\definecolor{mypink}{RGB}{144,71,145}
\definecolor{myred}{RGB}{143,4,7}
\definecolor{mygreen}{RGB}{0,128,0}
\definecolor{mydarkblue}{RGB}{6,57,112}
\definecolor{mylightblue}{RGB}{3,165,252}
\definecolor{hlightyellow}{RGB}{255,255,204}
\definecolor{lightyellow}{RGB}{255,255,204}
\definecolor{white}{RGB}{255,255,255}

\DeclareCaptionStyle{ruled}{labelfont=normalfont,labelsep=colon,strut=off}
\floatstyle{ruled}
\newfloat{listing}{tb}{lst}{}
\floatname{listing}{Listing}

\newif\ifanonsubmission

\makeatletter
\let\AAAI@orig@affiliations\affiliations
\newcommand{\AAAI@saved@affils}{}%
\renewcommand{\affiliations}[1]{%
  \gdef\AAAI@saved@affils{#1}
  \AAAI@orig@affiliations{#1}
}
\makeatother

\title{JRDB-Reasoning: A Difficulty-Graded Benchmark for Visual Reasoning in Robotics}

\author{
    Simindokht Jahangard\textsuperscript{\rm 1},
    Mehrzad Mohammadi\textsuperscript{\rm 2},
    Yi Shen\textsuperscript{\rm 1},
    Zhixi Cai\textsuperscript{\rm 1},
    Hamid Rezatofighi\textsuperscript{\rm 1}
}

\affiliations{
    \textsuperscript{\rm 1}Monash University\\
    \textsuperscript{\rm 2}Sharif University of Technology\\
   simindokht.jahangard@monash.edu

}

\begin{document}

\makeatletter
\begin{strip}
  \centering
  {\LARGE \@title \par}
  \vspace{1em}
  \ifanonsubmission
    {\large \textbf{Anonymous Authors} \par}
    \vspace{0.5em}
    {\small \textit{Affiliations withheld for double-blind review} \par}
  \else
    {\large \@author \par}
    \vspace{0.5em}
    {\small \AAAI@saved@affils \par} 
  \fi
  \vspace{1.25em}
  \includegraphics[width=\textwidth]{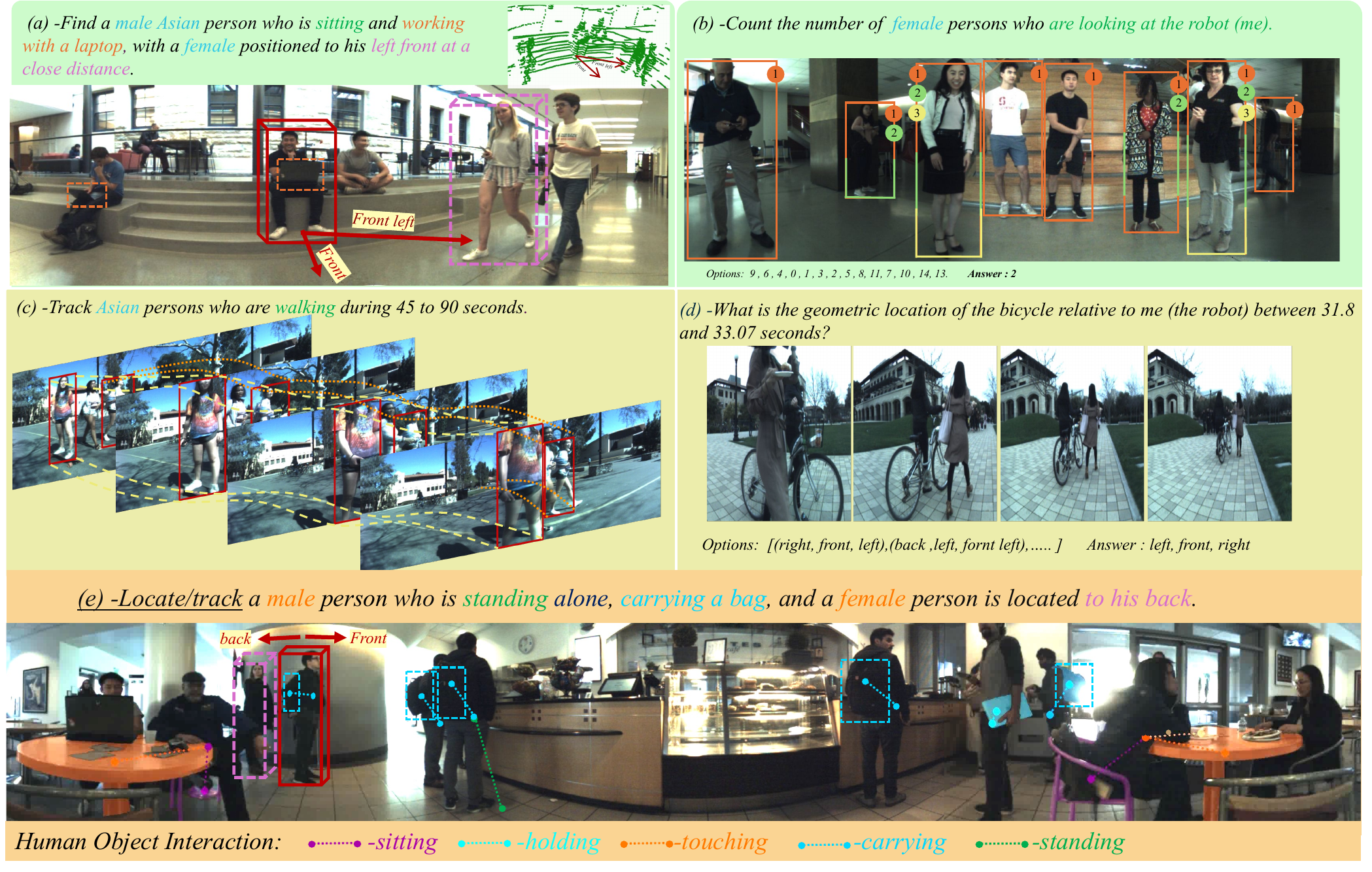}
  \vspace{-1.25em}
  \captionof{figure}{Illustration of JRDB-Reasoning on VG and VQA for images and videos.
    \textbf{(a)} and \textbf{(e)} depict VG tasks on images \& HOI.
    \textbf{(b)} presents a VQA-Counting question with reasoning steps:
    (1) identify all humans, (2) filter for females, (3) find female looking at the robot.
    \textbf{(c)} illustrates a VG multi-object tracking task,
    while \textbf{(d)} shows a VQA-\emph{Wh} question applied to a video.}
  \label{fig:teaser}
\end{strip}
\makeatother

\input{sec/0_abstract}
\input{sec/1_intro}

\input{sec/2_related_works}
\input{sec/3_Dataset}
\input{sec/4_experiments}
\input{sec/4-4-pic}
\input{sec/5_conclusion}

\bibliography{main}

\end{document}

%% file: sec/0_abstract.tex
\begin{abstract}
\vspace{-0.5em}
Recent advances in  Vision-Language Models (VLMs) and large language models (LLMs) have greatly enhanced visual reasoning, a key capability for embodied AI agents like robots. However, existing visual reasoning benchmarks often suffer from several limitations: they lack a clear definition of reasoning complexity, offer have no control to generate questions over varying difficulty and task customization, and fail to provide structured, step-by-step reasoning annotations (workflows). To bridge these gaps, we formalize reasoning complexity, introduce an adaptive query engine that generates customizable questions of varying complexity with detailed intermediate annotations, and extend the JRDB dataset with human-object interaction and geometric relationship annotations to create JRDB-Reasoning, a benchmark tailored for visual reasoning in human-crowded environments. Our engine and benchmark enable fine-grained evaluation of visual reasoning frameworks and dynamic assessment of visual-language models across reasoning levels.
\end{abstract}

%% file: sec/1_intro.tex
\section{Introduction}
Recent advances in Vision-Language Models (VLMs) and large language models (LLMs) have significantly transformed computer vision, evolving it from basic visual perception to higher-level visual reasoning~\cite{amizadeh2020neuro}. By integrating cognitive capabilities that closely resemble human perception and understanding, visual reasoning is especially crucial for embodied AI agents such as robots. These advances enable robots not only to perceive and interpret visual information but also to engage in cognition and reasoning—allowing them to make decisions, solve problems, and draw meaningful conclusions.

Developing models with advanced visual reasoning capabilities requires high-quality benchmarks that not only provide rigorous evaluation but also challenge models with diverse, context-rich, and complex reasoning tasks~\cite{wu2024star, patraucean2023perception, mangalam2023egoschema, chandrasegaran2024hourvideo}. The evolution from static 2D images~\cite{Yu_Poirson_Yang_Berg_Berg_2016, chen2024revisiting, johnson2017clevr} to dynamic videos, 3D environments, and multimodal datasets~\cite{li2024mvbench, shen2024longvu, fu2025video} has enabled models to capture temporal dependencies, integrate multimodal cues, and engage in high-level reasoning for real-world decision-making.
However, existing visual reasoning benchmarks often group under the broad label of “reasoning,” without distinguishing between different levels of complexity. In reality, visual reasoning spans a continuum—from simple perceptual tasks, such as object recognition, to multi-step logical deductions that require sequential and causal reasoning. This lack of granularity causes benchmarks to evaluate basic recognition tasks in the same way as those requiring complex, multi-stage inference. Consequently, current benchmarks are not able to accurately measure a model’s capacity for advanced reasoning, limiting their effectiveness in evaluating true reasoning capabilities.

Moreover, current datasets lack the ability to adjust question complexity based on user-specified parameters such as the type of task (\eg, VG, VQA), the focus of the question (\eg, humans, objects), and the level of spatial or temporal reasoning (\eg, in images or videos). This limitation hinders the customization of reasoning tasks for diverse evaluation needs, thereby restricting comprehensive model assessment and reducing adaptability to user-specific requirements.

Additionally, these datasets generally provide only input data and corresponding labels, while omitting the intermediate reasoning steps necessary to derive the final answer. Such steps are vital for evaluating models that rely on chain-of-thought (step-by-step) reasoning, such as workflow-based systems, neuro-symbolic frameworks, and program-guided models~\cite{suris2023vipergpt,ke2024hydra}, where understanding the reasoning trajectory is key to assessing the model’s inference process.

In this paper, we address all these gaps by: (i) defining and formalizing the complexity of perception-to-reasoning tasks, (ii) developing an adaptive query engine that dynamically generates non-predefined questions with adjustable complexity levels, (iii) providing step-by-step solutions (intermediate annotation) for each visual reasoning question within the workflow, and (iv) annotating human-object interactions manually and computing geometric relationships to enhance the robotic JRDB dataset for integration with an adaptive query engine.

First, we formalize the reasoning complexity of a visual reasoning task in terms of the number of reasoning steps required to reach a solution. By representing a scene and its entities as a spatio-temporal graph, we argue that the complexity of a visual reasoning question is closely tied to the number of nodes (entities, S) and edges (relationships, R), as well as their temporal dynamics (time slots, T). As the number of these factors increases, the overall task complexity rises. This structured formulation provides a systematic framework for assessing reasoning difficulty and evaluating model capabilities with greater precision.

Second, we introduce an adaptive query engine that dynamically generates non-predefined questions with adjustable complexity. Users can customize queries across spatial and temporal dimensions in images and videos, tailoring them to specific tasks, including task type (VG or VQA), modality (image or video), subject focus (human, object, or both), and spatial/temporal scope (single, pair, or clique).

Third, the engine generates step-by-step intermediate solutions for each query, providing workflow annotations that capture the entire reasoning trajectory. For example, as illustrated in Figure \ref{fig:teaser}(b), given the query “Count the number of female persons looking at the robot (me),” the engine produces a structured reasoning sequence: (1) identifying all humans, (2) filtering female individuals, and (3) detecting females looking at the robot. This decomposition enables fine-grained evaluation of reasoning accuracy and facilitates deeper analysis of VLMs or workflow-based framework, including compositional models~\cite{suris2023vipergpt,you2023idealgpt,ke2024hydra,cai2025naver}.

Fourth, we manually annotated human-object interactions with associated confidence levels, Figure \ref{fig:teaser}(e). In addition, we extracted geometric relationships, including spatial relations and distances between each entity (human or object). Through this process, we enriched the JRDB dataset~\cite{martin2021jrdb}, complementing and extending existing annotations~\cite{ehsanpour2022jrdb, jahangard2024jrdb, vendrow2023jrdb, le2024jrdb}. All annotations are integrated into our adaptive query engine to generate questions.

Finally, leveraging advances in VLMs~\cite{zhang2023video, wu2023next}, we assess their performance on JRDB-Reasoning, our novel visual reasoning benchmark, to evaluate their capacity for tasks of varying complexity. Despite the engine’s ability to generate highly complex questions, we focus on a simpler subset aligned with current model capabilities. These evaluations provide key insights into model performance, limitations, and the benchmark’s potential to drive future research.

%% file: sec/2_related_works.tex
\section{Related Works} 
\label{sec:related_works}
\textbf{Visual Reasoning Benchmarks.}
High-quality datasets and benchmarks drive visual reasoning progress, evolving from 2D to 3D and from images to videos, enabling models to tackle complex multimodal tasks like VG and VQA with richer cues and contexts for improved decision-making.
\begin{figure*}[t]
\begin{center}
\scalebox{1}{
 \includegraphics[width=1\linewidth]{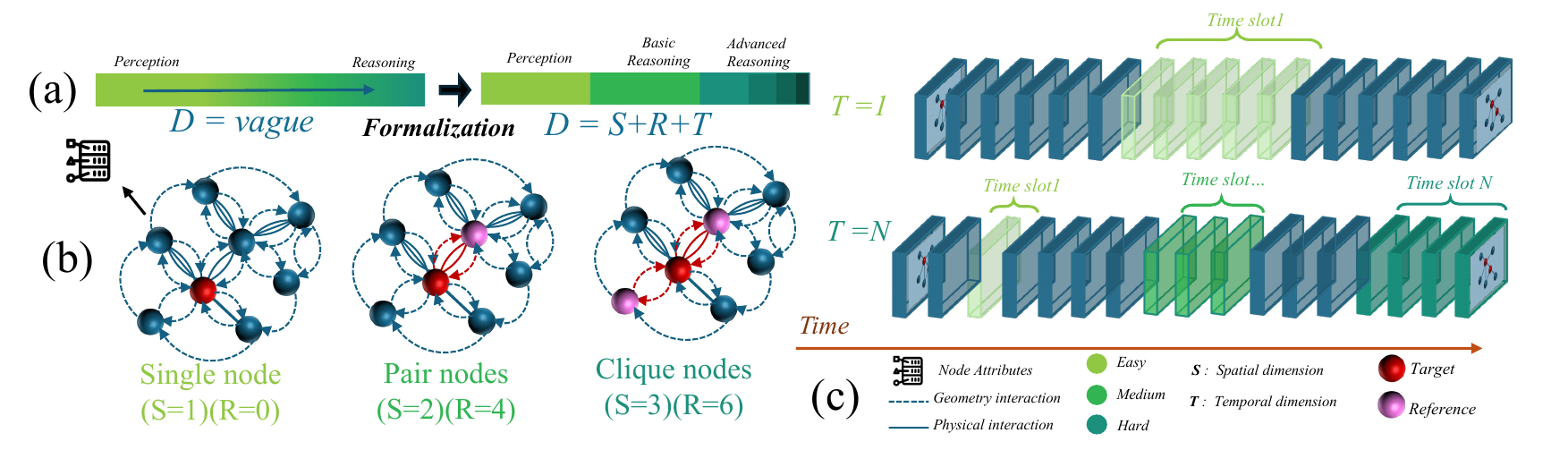}
}
\end{center}
\vspace{-1em}
  \caption{Formalization of complexity involves nodes (S), interactions (R), and time slots (T); greater involvement indicates higher complexity (darker green), while lighter green shows lower complexity}
\label{fig:node}
\vspace{-1.5em}
\end{figure*}
Benchmarks evolved from 2D images to videos and 3D environments.
Early benchmarks like Flickr30K~\cite{plummer2015flickr30k} and DAQUAR Visual7W~\cite{zhu2016visual7w} focused on entity localization, but lacked the scene context necessary for testing deeper reasoning abilities. Datasets like RefCOCO Series~\cite{kazemzadeh2014referitgame}, Ref-Adv~\cite{akula_words_2020}, COCO-QA~\cite{ren2015exploring}, Visual Genome~\cite{krishna2017visual}, VQAv2~\cite{goyal2017making}, CLEVR~\cite{johnson2017clevr} and GQA~\cite{hudson2019gqa} introduced more complex tasks, enhancing visual grounding and improving the ability to infer relationships and interactions in more context-rich environments. 
With video’s rise, benchmarks such as TGIF-QA~\cite{jang2017tgif}, TVQA~\cite{lei2018tvqa}, STAR~\cite{wu2024star}, VideoQA~\cite{yang2003videoqa}, MVBench~\cite{li2024mvbench}, Perception Test~\cite{patraucean2023perception}, VITATECS~\cite{li2023vitatecs}, LLAVIDAL~\cite{chakraborty2024llavidal}, MLVU~\cite{zhou2024mlvu}, Towards Event-oriented~\cite{du2024towards}, and TempCompass~\cite{liu2024tempcompass} advanced temporal reasoning in dynamic, event-based, and egocentric contexts. Datasets like OSCaR~\cite{nguyen2024oscar}, EgoSchema~\cite{mangalam2023egoschema}, and HourVideo~\cite{chandrasegaran2024hourvideo} further pushed first-person perspective research in interactive reasoning, action tasks, and large-scale egocentric video understanding.
Furthermore, 3D benchmarks like ScanReason~\cite{zhu2024scanreason},
 LAMM~\cite{yin2023lamm},
SpatialRGPT~\cite{cheng2025spatialrgpt},
Mono3DVG~\cite{zhan2024mono3dvg}, and M3DBench~\cite{li2023m3dbench}, and in autonomous driving like  ScanQA-3D~\cite{azuma2022scanqa}, NuScenes-QA~\cite{qian2024nuscenes}  focus on spatial reasoning in realistic 3D environments, addressing the need for models that can navigate and understand immersive spaces. while visual reasoning benchmarks expand the scope to include dynamic and interactive environments that test models on adaptability and complex reasoning in real-world applications.
This expansion into more complex, multimodal, and spatially aware datasets has enabled models to address increasingly dynamic and interactive challenges. 
Visual reasoning benchmarks lack clear definitions of difficulty levels, customization of question complexity, and reasoning steps, hindering accurate model evaluation. In this paper, we define perception-to-reasoning complexity, build an adaptive query engine for dynamic questions with adjustable difficulty, provide step-by-step solutions, and introduce JRDB-Reasoning, a dataset for robots in human-crowded environments.
\\
\textbf{Vision Language Models.} 
Multimodal models, particularly VLMs, have advanced from perception to complex reasoning, with increasing emphasis on reasoning capabilities, as reviewed here. 
These advances build on LLM progress in language and reasoning, as seen in GPT-4~\cite{achiam2023gpt}, Gemini~\cite{team2023gemini}, Claude, and GPT-4o~\cite{hurst2024gpt}. Meanwhile, models such as VideoChat~\cite{li2023videochat}, Visual ChatGPT~\cite{wu2023visual}, VALLEY~\cite{luo2023valley}, Otter~\cite{li2023otter}, and MiniGPT-4~\cite{zhu2023minigpt}, along with newer methods like LLaVA-NeXT-Video~\cite{zhang2024llava} and LongVU~\cite{shen2024longvu}, have focused on advancements in video understanding and visual processing. Additional efforts from models like mPLUG-Owl2~\cite{ye2024mplug}, SPHINX~\cite{lin2023sphinx}, Intern-VL~\cite{chen2024internvl}, Yi-VL~\cite{ai2024yi}, VideoChat2~\cite{li2023videochat}, Cambrian-1~\cite{tong2024cambrian}, PLLaVA~\cite{xu2024pllava}, Blip2~\cite{li2023blip}, Florence2~\cite{xiao_florence-2_2024}, and MiniGPT4-Video~\cite{ataallah2024minigpt4}, as well as recent models such as Emu3~\cite{wang2024emu3} and Pixtral~\cite{agrawal2024pixtral}, have significantly contributed to improving video comprehension, contextual learning, and instruction tuning. Notably, models like GLIP~\cite{li2022grounded}, ReCLIP~\cite{subramanian2022reclip},  GroundingDINO~\cite{chen2022grounding}, and YOLO-World~\cite{cheng_yolo-world_2024}, Gemini~\cite{team2023gemini}, GPT-4o~\cite{hurst2024gpt}, Qwen2-VL~\cite{wang2024qwen2}, LLaVA-Video~\cite{zhang2024llava}, InternVL2~\cite{chen2024internvl}, LongVU~\cite{shen2024longvu}, and Flash-VStream~\cite{zhang2024flash} advance more in terms of multimodal understanding, long-context reasoning, and efficient video and language processing.
 Together, these advances highlight the growing potential of multimodal methods to process and integrate diverse forms of information.
However, the ability of these methods to handle reasoning at different levels of difficulty remains unexplored. In this paper, we evaluate some of these models' reasoning performance across various difficulty levels using the JRDB-Reasoning to analyze their capacity for reasoning in complex visual scenarios.

%% file: sec/3_Dataset.tex
\section{Visual Reasoning Engine and Dataset}
\subsection{Formalizing Reasoning Complexity}
Reasoning tasks vary in complexity, ranging from simple object recognition to more intricate problems requiring multiple reasoning steps. As the number of steps increases, the difficulty correspondingly rises. We use this hypothesis as a foundation to formalize visual reasoning complexity here.

Figure~\ref{fig:node} provides an overview of our approach to formalizing reasoning complexity in visual reasoning. We propose that perception and reasoning exist on a continuum (difficulty is vague) rather than as distinct levels. As illustrated in Figure~\ref{fig:node}(a), reasoning complexity spans from basic perception to advanced multi-step reasoning, typically depending on the number of intermediate steps required to reach a solution. But, this complexity cannot always be quantified before solving a visual reasoning question.

To address this, we propose an alternative complexity measure that strongly correlates with the number of reasoning steps. Specifically, we represent a scene as a dynamic spatio-temporal graph, where nodes correspond to entities such as humans, objects, and surfaces, while edges capture various types of interactions (\eg geometric, physical), as shown in Figure~\ref{fig:node}(b). It can be a target (question asked about it) or a reference (related to the target). Each entity (S) is a unique instance (\eg, a person, an object, or a surface) tracked over time. These entities and their interactions (R) evolve across different time intervals, referred to as time slots (T), as illustrated in Figure~\ref{fig:node}(c). 

As more nodes, edges, and time slots are introduced, reasoning complexity in a visual reasoning question increases, indicated by the color transition from light green to dark green in Figure~\ref{fig:node}. This structured approach provides a systematic framework for understanding reasoning complexity, enabling a more precise evaluation of model capabilities.

\begin{figure*}[!t]
\begin{center}
\scalebox{1}{
\includegraphics[width=1\linewidth]{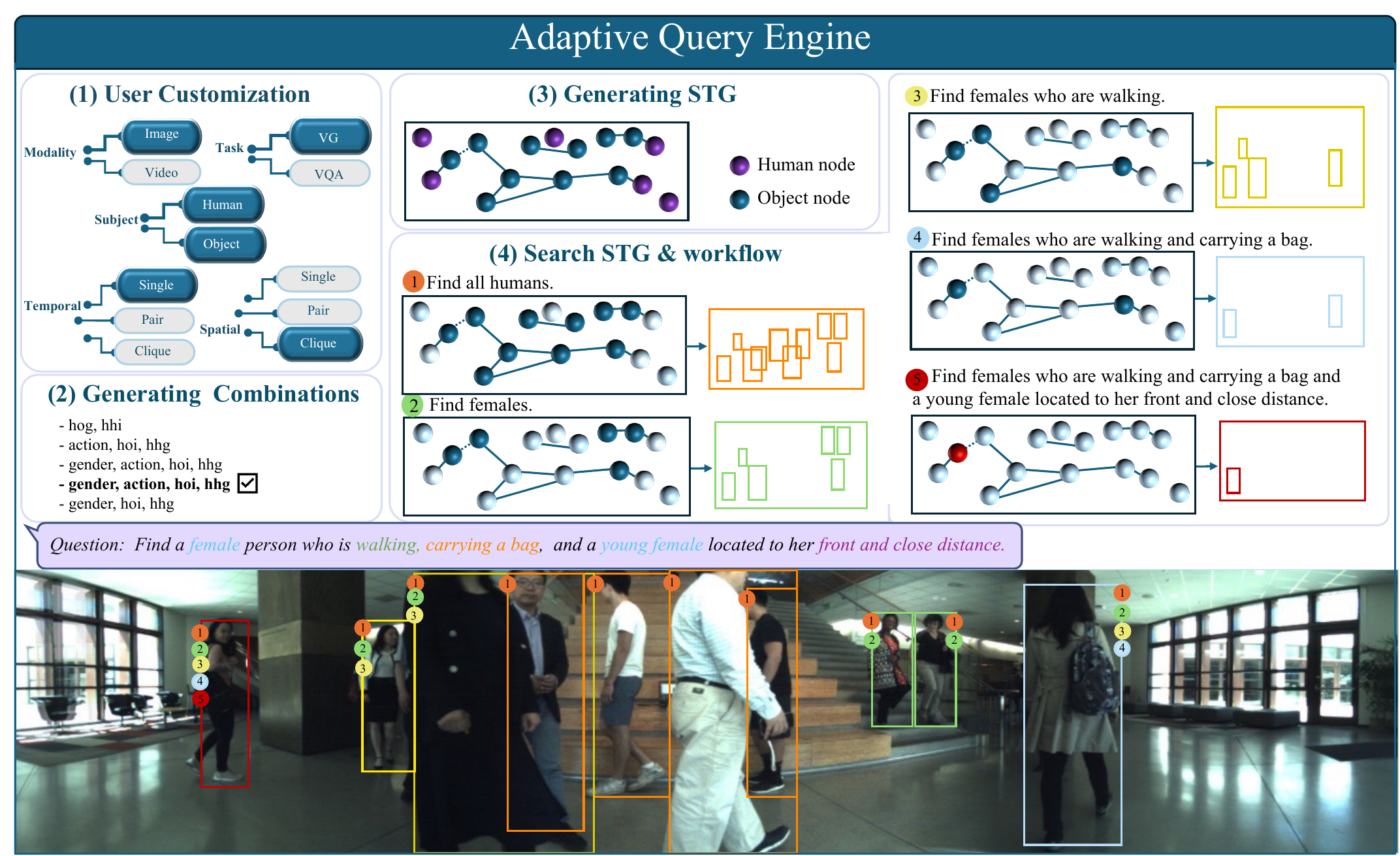}}
\end{center}
\vspace{-1em}
\caption{\textbf{Example of the Generative Query-Based Engine:} (1) user selects preferences: modality (image), subject (human\&object), task (VG), spatial (clique), temporal (single). (2) All attribute combinations are computed. (3) a spatial-temporal graph (STG) is generated. (4) The STG is searched to generate intermediate annotations and queries.}
\label{fig:Engine}
\vspace{-1.5em}
\end{figure*}
To estimate the complexity of a visual reasoning question, we introduce formulas modeling element involvement, spatial interactions, and temporal spans in the dynamic scene graphs.
To validate this formulation, we conducted a user study with 20 participants who rated the difficulty of 500 questions. The results showed strong agreement between human judgments and our computed scores, confirming that questions involving more entities, spatial reasoning, or temporal tracking are perceived as more complex.

\textbf{(1) Entity (Node) Involvement (S).} Question difficulty depends first on how many entities (or nodes) are involved, target or reference in Figure~\ref{fig:node}. This is quantified by counting the nodes \( N = \{n_1,\dots, n_k\} \), where \( k \) is the total number of entities. Each node \( n_i \) has a unique ID and set of attributes, which can be static (e.g., age, gender) or dynamic (e.g., actions, states). Entity involvement \( S \) is expressed as 
$S = k$.

\textbf{(2) Spatial Relationships (R).}  
This factor captures spatial interactions between entities within a single frame, represented by edges \( E_s \) connecting nodes \( n_i, n_j \in N \), i.e.,
\[
E_s = \{(n_i, n_j) \mid n_i, n_j \in N\}.
\]
These edges encode properties like proximity or orientation. The spatial interaction complexity is defined as
\[
R = |E_s|,
\]
where \( |E_s| \) denotes the number of spatial relationships (edges) at a given time frame.

\textbf{(3) Temporal Interactions (T).}  
This factor captures cross-temporal interactions via edges \( E_t \) connecting entities \( n_i, n_j \in N \) over time intervals \( t_1, t_2 \), i.e.,
\[
E_t = \{(n_i, n_j, t_1, t_2) \mid n_i, n_j \in N,\ t_1, t_2 \in \text{time frames}\}.
\]
These edges capture continuity or tracking across time. The temporal interaction complexity is defined as
\[
T = |E_t|,
\]
where \( |E_t| \) is the number of time slots involving each entity.

\textbf{Overall Difficulty (D).}  
The difficulty of a visual question depends on entity involvement (\(S\)), spatial interactions (\(R\)), and temporal interactions (\(T\)), $D = S+ R + T$.

By adopting this approach, we develop a more systematic framework for understanding task difficulty, which allows for a nuanced evaluation of model performance. This structured methodology not only improves benchmarking techniques but also offers deeper insights into a model's capability to progress from basic perception to advanced reasoning.
\subsection{Generative Adaptive Query Engine}
We developed an adaptive query engine that enables users to generate open-ended, compositional questions with customizable complexity, based on their preferences across spatial and temporal scales in images or videos. It supports diverse tasks such as VG and VQA (Wh and Counting type question, Figure \ref{fig:teaser}(b)(d)). In addition, the engine produces intermediate annotations that provide detailed descriptions at every stage of reasoning. These annotations offer deeper interpretability and are particularly valuable for evaluating VLMs that rely on multi-step reasoning and compositional understanding.

The engine operates based on user-defined preferences, including task type (VG or VQA), modality (image or video), subject focus (human, object, or both), and spatial/temporal scope (single, pair, or clique). Depending on the selected modality, a spatial-temporal graph is constructed. Based on the spatial configuration, all possible combinations of node attributes and edge relationships are computed, and one combination is randomly selected. The search process then proceeds over the spatial-temporal graph using this chosen combination. At each step, corresponding reasoning steps—such as intermediate annotations—are generated, and the selected attributes or relationships are incorporated into the query. Further details of the generative query engine, along with an illustrative example, are provided below.
As shown in Figure~\ref{fig:Engine}, the proposed engine operates through the following steps:
(1) Users begin by customizing their query preferences via parameters that define the difficulty level. In this example, the user selects the VG task, specifies both human and object as subjects, chooses the image modality ($T = 1$), and sets the spatial preference to a clique configuration with three nodes ($S=3$).
(2) Given the spatial configuration, all possible combinations of node attributes and edge relationships are computed, and one combination is randomly selected. For instance, the attribute set [“gender”, “action”, “hoi”, “hhg”] with ($R = 2$) may be chosen, where “hoi” stands for human-object interaction, and “hhg” refers to human-human geometry.
(3) A spatial-temporal graph (STG) is then generated based on the selected modality.
(4) Finally, the engine searches through the spatial-temporal graph and incrementally generates both intermediate annotations and the corresponding questions. See Supplementary Material for used templates.
\subsection{Dataset and Annotations}
\begin{figure*}
\begin{center}
\scalebox{1}{
 \includegraphics[width=1\linewidth]{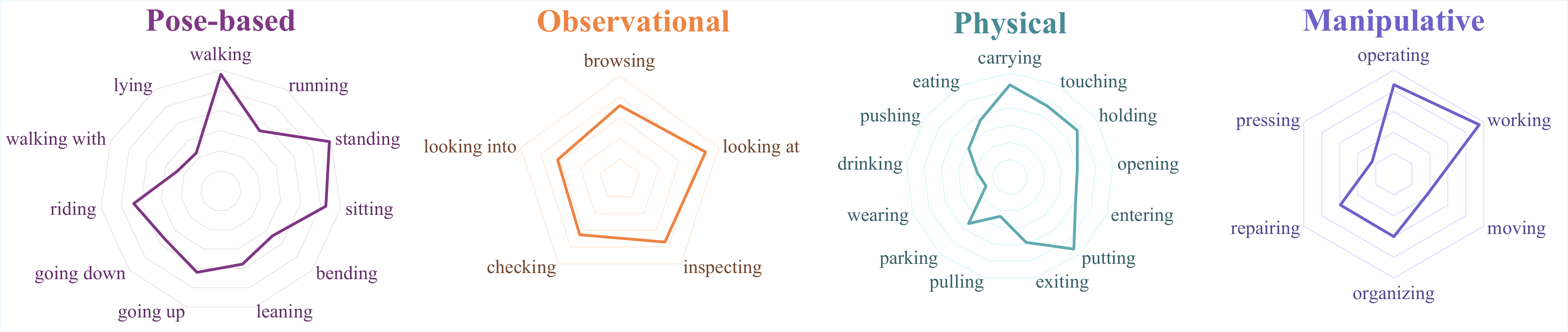}}
\end{center}
\vspace{-1.5em}
  \caption{Human-object interaction falls into four types: Pose-based, Observational, Physical, and Manipulative.}
\label{fig:HOI}
\vspace{-1.5em}
\end{figure*}

\textbf{Human Object Interaction.}
Human-object interaction (HOI) enhances visual reasoning by connecting human actions to objects, improving identification, activity recognition, and scene interpretation. It also extends the range of questions generated in JRDB annotations. In JRDB-Reasoning, we offer multi-label, fine-grained HOI annotations,shown in Figure \ref{fig:teaser}(e), at the frame level, categorized into four groups, Figure~\ref{fig:HOI}. The first category focuses on pose-based activities such as walking, standing, and sitting,shown by dash Figure \ref{fig:teaser}(e), which are the most common in the dataset. The second category involves observational interactions, where individuals visually engage with objects or their surroundings, but these are less frequent. The third category emphasizes physical interactions with objects, like touching and carrying, and is strongly represented. The fourth category includes manipulative interactions, involving precise handling of objects, such as operating and working, which are also notable. These categories provide a detailed framework for analyzing human-object interactions in dynamic environments, with pose-based and physical interactions being the most common, while observational interactions are the least frequent.

\noindent\textbf{Annotation Process.}
To annotate human-object interactions, we use a specialized toolbox that assigns unique IDs to 2D and 3D bounding box annotations. Our annotation process follows a quality control procedure aligned with previous JRDB benchmarks to ensure consistency. Interaction annotations are carefully aligned with individual actions, and strict guidelines are followed to maintain accuracy. We also address challenges such as robot distance, varying lighting conditions, occlusions, and crowded environments by assigning difficulty levels—Easy (1), Medium (2), and Hard (3)—based on annotator confidence. Each label is reviewed by two additional annotators, with random quality checks to enhance fairness and reliability. More information are provided in Supplementary material.\\
\noindent\textbf{Geometry Relationships.} 
Geometry relationships, including distance and spatial relationship , improve visual reasoning by providing context for object recognition, navigation, and decision-making. They help systems understand object positioning, track movement, predict interactions, and optimize path finding. In robotics and computer vision, this enhances scene understanding, human-robot interaction, and autonomous decision-making.\\
\noindent\textbf{Spatial relationships.} 
To extract spatial geometry relationships, inspired by ~\cite{qian2024nuscenes}, we compute the spatial relationship relative to a target point, which could be a robot or a person with orientation. The target point serves as the origin for measuring distances and angles in 3D space. As explained in JRDB~\cite{martin2021jrdb}, a robot has a fixed orientation, while for humans, their orientation is also taken into account, shown with red arrow in Figure \ref{fig:teaser}(e). This captures spatial relationships by accounting for both position and facing direction, yielding a more accurate environmental representation.
To achieve this, we compute the angle between the vector that connects the centers of two bounding boxes and the forward direction of target. The formula for this angle \( \theta \) is expressed as:
\[
\theta = \cos^{-1} \left( \frac{(B_1[:2] - B_2[:2])}{\| B_1[:2] - B_2[:2] \|} \right)
\]
where 
  \( B_i = [x, y] \) represents the 3D bounding box of entity \( i \).
Here, the forward direction of the target point (e.g., robot or human) is defined as \( 0^\circ \), with counterclockwise angles being considered positive. For spatial relationships, we calculate the relative position between two points, \( B_1 \) and \( B_2 \), along both the X and Y coordinates. Additionally, we take the Z-coordinate into account and calculate the distance, incorporating this third dimension (up or down) into the analysis which resulted in 16 possible directions.
Based on the calculated angle, the relationship between the two objects is categorized in Table~\ref{table:condition_theta_distance_split}.\\
\noindent\textbf{Distance.} 
To enhance interpretability and streamline analysis, we divide distances into five categories based on predefined thresholds (see Table~\ref{table:condition_theta_distance_split}). This classification offers a clear and consistent framework for analyzing distance measures. For human-human interactions, distances are computed between their 3D bounding boxes~\cite{martin2021jrdb}. In human-object scenarios, we use point cloud data and object annotations~\cite{le2024jrdb}, applying efficient nearest neighbor searches via \textit{KD-trees}\footnote{\url{https://en.wikipedia.org/wiki/K-d_tree}}. A KD-tree is built for each point cloud pair to perform k-nearest neighbor queries, and the mean neighbor distance quantifies proximity. See supplement for implementation details.
\begin{table}[b]
\vspace{-1.5em}
   \centering
   \scalebox{0.7}{
   \begin{tabular}{l|c||l|c}
    \toprule
    \textbf{Relation} & \textbf{Condition on \(\theta\)} & \textbf{Distance} & \textbf{Condition on \(d\)} \\
    \hline
    \text{(up/down) front}        & \(-30^\circ \leq \theta \leq 30^\circ\)   & Very close & \(d < 0.5\) \\
    \text{(up/down) front left}   & \(30^\circ < \theta \leq 45^\circ\)        & Close      & \(0.5 \leq d < 1.5\) \\
    \text{(up/down) left}         & \(45^\circ < \theta \leq 135^\circ\)       & Moderate   & \(1.5 \leq d < 5\) \\
    \text{(up/down) back left}    & \(135^\circ < \theta \leq 145^\circ\)      & Far        & \(5 \leq d < 10\) \\
    \text{(up/down) front right}  & \(-45^\circ \leq \theta < -30^\circ\)       & Very far   & \(d \geq 10\) \\
    \text{(up/down) right}        & \(-135^\circ \leq \theta < -45^\circ\)      &            &  \\
    \text{(up/down) back right}   & \(-145^\circ \leq \theta < -135^\circ\)     &            &  \\
    \text{(up/down) back}         & \text{else}                                &            &  \\
    \hline
   \end{tabular}}
\vspace{-0.5em}
\caption{Spatial geometric relationship based on \(\theta\) \& \(d\)}
\label{table:condition_theta_distance_split}
\vspace{-2em}
\end{table}

%% file: sec/4_experiments.tex
\section{Experiment}
\begin{table}[b]
\vspace{-1.5em}
   \centering
   \footnotesize
   \begin{tabular}{p{2.2cm} c c c}
    \toprule
  \multirow{2}{*}{{\bf Multi-modal LLM}} 
  & {\bf $\mathcal{D}_1$} & {\bf $\mathcal{D}_2$} & {\bf $\mathcal{D}_3$}\\
\cmidrule(lr){2-4}
& {\tiny (S= T= 1, R= 0)} & {\tiny (S= 2, R $\ge$ 1, T= 1)} & {\tiny (S= 3, R $\ge$ 2, T= 1)}  \\
    \midrule
    InternVL 2.5& \textbf{49.9} & \textbf{49.2} & \textbf{44.6}  \\
    Paligemma &  48.6  &  47.5 & 44.6 \\
    Qwen2.5-VL & 36.3 & 33.0 &  27.7 \\
    LLaVA-NeXT &  35.4 & 33.2 & 28.7 \\
    \midrule
    Human Eval & 96.8 & 88.5 & 87.1\\
    \bottomrule
  \end{tabular}
\vspace{-1em}
\caption{{\bf VQA Experiment}: VLM evaluation on images of varying difficulty levels using accuracy on JRDB-Reasoning}
\label{table:VQA-image}
\vspace{-1em}
\end{table}
\begin{table}[t]  
   \centering
   \footnotesize
   \begin{tabular}{p{2.2cm} c c c}
    \toprule
   \multirow{2}{*}{{\bf Multi-modal LLM}} 
  & {\bf $\mathcal{D}_1$} & {\bf $\mathcal{D}_2$} & {\bf $\mathcal{D}_3$}\\
\cmidrule(lr){2-4}
& {\tiny (S= T= 1, R= 0)} & {\tiny (S= 2, R $\ge$ 1, T= 1)} & {\tiny (S= 3, R $\ge$ 2, T= 1)}  \\
    \midrule
    InternVL 2.5 &  \textbf{49.1}  & \textbf{48.6} &  \textbf{46.0} \\
    Qwen2.5-VL & 47.3 & 41.8 & 36.5  \\
    \mbox{LLAVA-NeXT-Video}&  41.2& 36.8 & 29.71 \\
    \midrule
    Human Eval & 98.3 & 92.1 & 89.1 \\
    \bottomrule
  \end{tabular}
\vspace{-1em}
\caption{{\bf VQA Experiment}: VLM evaluation on videos of varying difficulty levels using accuracy on JRDB-Reasoning}
\label{table:VQA-video}
\vspace{-2em}
\end{table}
With the rapid advancement of VLMs and multi-modal architectures capable of integrating heterogeneous modalities with increasingly sophisticated reasoning abilities, our work aims to provide a systematic and rigorous evaluation of their performance across multiple levels of task complexity. In particular, we focus on visual reasoning tasks, encompassing both VG and VQA, to examine the models’ reasoning and comprehension capabilities in multi-modal settings. To complement automated evaluation metrics, we conduct a human evaluation with three expert annotators, reporting the mean of their scores to ensure robustness and reliability. Furthermore, we introduce a standardized evaluation toolkit, carefully designed to maintain consistent difficulty levels and enable a fair comparative analysis across different VLMs.

\noindent\textbf{Evaluation Metrics.} For VG, we adopt the evaluation metric from prior research ~\cite{suris2023vipergpt, ke2024hydra, cai2025naver}, where a predicted bounding box is considered correct if its mIoU with the ground-truth box exceeds 0.5. For VQA, we use the accuracy metric as described in ~\cite{johnson2017clevr, jahangard2024jrdb, hudson2019gqa}.

\noindent\textbf{Setup and Details.} 
To evaluate the models, we utilize the JRDB-Reasoning dataset, which contains questions of varying difficulty. We categorize the dataset into three difficulty levels, denoted by $\mathcal{D}_1$, $\mathcal{D}_2$, and $\mathcal{D}_3$, defined as 
{\small 
$\mathcal{D}_1 = \{ (S, R, T) \mid S = 1, R = 0, T = 1 \},\;
 \mathcal{D}_2 = \{ (S, R, T) \mid S = 2, R \geq 1, T = 1 \},\;
 \mathcal{D}_3 = \{ (S, R, T) \mid S = 3, R \geq 2, T = 1 \},$
}
where $S$ denotes the number of nodes (humans or objects), $R$ represents the number of interactions (edges), and $T$ indicates the number of time slots.
$\mathcal{D}_1$ includes questions about a single node ($S = 1$) and its attributes, with no interactions ($R = 0$) and a single time slot ($T = 1$). $\mathcal{D}_2$ increases complexity by involving two nodes ($S = 2$) and at least one interaction ($R \geq 1$), still with $T = 1$. $\mathcal{D}_3$ requires reasoning over three nodes ($S = 3$) and at least two interactions ($R \geq 2$), with $T = 1$. We exclude levels beyond $\mathcal{D}_3$ as current models struggle with higher complexities in image and video reasoning. 
To ensure a focused evaluation, we conduct experiments using a single time slot ($T = 1$). This setup aligns with the design of many existing models, such as video grounding methods, which are typically developed for single-object tracking rather than multi-object temporal interactions.\\
\begin{table}[b]
\vspace{-1.5em}
\centering
\footnotesize
\renewcommand{\arraystretch}{0.9} 
\begin{tabular}{p{2.2cm} c c c}
\toprule
\multirow{2}{*}{{\bf Multi-modal LLM}} 
  & {\bf $\mathcal{D}_1$} & {\bf $\mathcal{D}_2$} & {\bf $\mathcal{D}_3$}\\
\cmidrule(lr){2-4}
& {\tiny (S= T= 1, R= 0)} & {\tiny (S= 2, R $\ge$ 1, T= 1)} & {\tiny (S= 3, R $\ge$ 2, T= 1)}  \\
\midrule
Florance-V2   & \textbf{37.6} & \textbf{35.7} & \textbf{25.2}\\
GroundingDINO & 25.8          & 22.5          & 14.7 \\
Qwen2.5-VL    & 19.8          & 17.4          & 14.0 \\
YOLO-World    & 13.5          & 5.3           & 3.1 \\
\midrule
Human Eval    & 95.5          & 87.6          & 86.6 \\
\bottomrule
\end{tabular}
\vspace{-1em}
\caption{{\bf VG Experiment}: VLM evaluation on images of varying difficulty levels using mIoU on JRDB-Reasoning}
\label{table:VG}
\vspace{-1.5em}
\end{table}
\noindent\textbf{Visual Grounding.} We selected several advanced models and evaluated them in zero-shot model, including GroundingDINO~\cite{liu2024grounding}, Florence-v2~\cite{xiao2023florence}, YOLO-World~\cite{cheng2024yolo}, and Qwen2.5-VL~\cite{bai2025qwen2}. As shown in Table~\ref{table:VG}, we observe that as the value of $D$ increases—indicating more nodes (entities) and connections—the complexity of the questions also rises, leading to a decline in model performance. 
Among the models, Florence-V2, Qwen2.5-VL and Grounding DINO consistently demonstrated the best and most stable performance at all difficulty levels. In contrast, YOLO-World dropped in performance with increased complexity, which could be due to their inability to handle more intricate or ambiguous image-text relationships, leading to a decline in performance as the tasks became more challenging.
The human evaluation row shows scores given by human evaluators to assess the performance of multi-modal LLMs. The scores (95.5, 87.6, and 86.6) reflect how well the models are perceived, with the highest score (95.5) being for $\mathcal{D}_1$ where it is single node ($S$ =1) and $T=1$ and there is no interaction, which likely involves the easiest questions. The decrease in scores with $\mathcal{D}_2$  and $\mathcal{D}_3$ may be due to added complexity in $R$, $R \geq 1$ and $R \geq 2$, more interactions such as geometric questions and challenges with stitched images, where the asked person being on either side of the image could make some questions harder for the evaluator.\\
\noindent\textbf{Step-by-Step Evaluation.}
We evaluate state-of-the-art models on the VG task using the JRDB-Reasoning benchmark, designed with progressively increasing compositional complexity. In $\mathcal{D}_1$, queries involve individual-level attributes—[``human'', ``gender'', ``age'', ``action'']—evaluated in four steps: detecting a person, then filtering by gender, age, and action (e.g., `` \emph{Find a young female person who is standing}''). $\mathcal{D}_2$ introduces relational reasoning to previous steps (e.g., `` \emph{...chatting with a man}''), while $\mathcal{D}_3$ incorporates spatial relations (e.g., `` \emph{...with a white man located to her right}''). As complexity increases, model performance degrades consistently , see Figure~\ref{fig:mean_iou_all_models}.
While all models perform well on basic detection and attribute filtering (steps 1–3), performance steadily declines as later steps introduce increasingly complex reasoning challenges. YOLO-World performs competitively at early stages but falters when relational and spatial comprehension is required. In contrast, models like Qwen 2.5-VL and Florence-2 exhibit a more gradual decline, demonstrating stronger integration of linguistic and visual understanding. Their ability to parse nuanced syntax and semantics allows more faithful alignment with complex human query.
More analysis provided in supplement.\\
\noindent\textbf{Visual Question Answering.}
We conduct VQA experiments on images using state-of-the-art models, including Paligemma~\cite{beyer2024paligemma}, 
Qwen2.5-VL~\cite{liu2024improved}, 
LLaVA-NeXT~\cite{liu2024improved} and InternVL 2.5~\cite{zhang2023video}, as shown in Table~\ref{table:VQA-image}. As seen in the results, the performance follows the same trend observed with VG, where as the complexity $D$ increases, model performance decreases. This is due to the increasing difficulty of the questions. Additionally, we observe that InternVL 2.5 performs better than the other models, likely due to its enhanced ability to handle complex image relationships, improving its understanding of object interactions, spatial reasoning, and context within the images. Qwen2.5-VL~\cite{bai2025qwen2}, InternVL 2.5~\cite{zhang2023video}, and LLaVA-NeXT-Video~\cite{zhang2024llava} are evaluated for video. Consistent with the trends seen in models evaluated on VG, performance drops as question complexity increases. However, model performance on video is better, as they can analyze multiple frames to answer the questions. In summary, model performance decreases with more complex questions, and they still underperform compared to humans.

%% file: sec/4-4-pic.tex
\colorlet{plot1color}{black}
\colorlet{plot2color}{teal}
\colorlet{plot3color}{violet}

\definecolor{FlorenceV2LineColor}{HTML}{1f77b4}
\definecolor{GroundingDINOLineColor}{HTML}{ff7f0e}
\definecolor{Qwen25VLLineColor}{HTML}{2ca02c}
\definecolor{YOLOWorldLineColor}{HTML}{d62728}

\pgfmathsetmacro{\commonxmin}{0.5}
\pgfmathsetmacro{\commonxmax}{6.5} 
\def\commonticks{1,2,3,4,5,6}    
\pgfmathsetmacro{\stepBaselineY}{2.5}  

\pgfplotsset{
    common axis style/.style={
        ylabel style={font=\scriptsize, align=center},
        xtick=\commonticks,
        xticklabels={}, 
        y tick label style={font=\scriptsize},
        ytick={0,20,40,60,80}, 
        grid=major,
        grid style={dashed, draw=gray!40},
        ymin=0,
        xmin=\commonxmin,
        xmax=\commonxmax, 
        width=\linewidth, 
        height=3.6cm,
        point meta=y,
    },
    florence2 plot style/.style={
        draw=FlorenceV2LineColor, mark=*,
        mark size=2pt, mark options={fill=white, line width=0.6pt},
        solid, line width=1pt
    },
    groundingdino plot style/.style={
        draw=GroundingDINOLineColor, mark=square*,
        mark size=2pt, mark options={fill=white, line width=0.6pt},
        dashed, line width=1pt
    },
    qwen25vl plot style/.style={
        draw=Qwen25VLLineColor, mark=triangle*,
        mark size=2pt, mark options={fill=white, line width=0.6pt},
        dash dot, line width=1pt
    },
    yoloworld plot style/.style={
        draw=YOLOWorldLineColor, mark=diamond*,
        mark size=2pt, mark options={fill=white, line width=0.6pt},
        dotted, line width=1pt
    },
}

\begin{figure}[t]
\setcounter{figure}{4} 
  \centering

  \begin{tikzpicture}
    \begin{axis}[
        common axis style,
        ylabel={($\mathcal{D}_1$)\\mIoU(\%)},
        ymax=85, 
        xmax=4.5, 
        extra description/.code={
            \draw[black, line width=1pt] (axis cs:\commonxmin,0) -- (axis cs:4.5,0); 
            \node[font=\scriptsize, color=plot1color, anchor=base] at (axis cs:1, \stepBaselineY) {Step$_1$};
            \node[font=\scriptsize, color=plot1color, anchor=base] at (axis cs:2, \stepBaselineY) {Step$_2$};
            \node[font=\scriptsize, color=plot1color, anchor=base] at (axis cs:3, \stepBaselineY) {Step$_3$};
            \node[font=\scriptsize, color=plot1color, anchor=base] at (axis cs:4, \stepBaselineY) {Step$_4$};
        }
    ]
    \addplot[florence2 plot style] coordinates {
        (1, 80.1) (2, 68.3) (3, 55.2) (4, 42.5)
    };
    \addplot[groundingdino plot style] coordinates {
        (1, 79.6) (2, 64.8) (3, 52.9) (4, 40.7)
    };
    \addplot[qwen25vl plot style] coordinates {
        (1, 78.3) (2, 63.5) (3, 51.9) (4, 44.8)
    };
    \addplot[yoloworld plot style] coordinates {
        (1, 78.1) (2, 51.3) (3, 32.1) (4, 21.3)
    };

    \node[font=\scriptsize, text=FlorenceV2LineColor, anchor=west] at (axis cs:3.4, 80) {
        \tikz[baseline=-0.5ex] \path[mark=*, mark size=2pt, mark options={fill=white, line width=0.6pt, draw=FlorenceV2LineColor}] plot coordinates {(0,0)};
        Florence-V2
    };
    \node[font=\scriptsize, text=GroundingDINOLineColor, anchor=west] at (axis cs:3.4, 72) {
        \tikz[baseline=-0.5ex] \path[mark=square*, mark size=2pt, mark options={fill=white, line width=0.6pt, draw=GroundingDINOLineColor}] plot coordinates {(0,0)};
        G-DINO
    };
    \node[font=\scriptsize, text=Qwen25VLLineColor, anchor=west] at (axis cs:3.4, 62) {
        \tikz[baseline=-0.5ex] \path[mark=triangle*, mark size=2pt, mark options={fill=white, line width=0.6pt, draw=Qwen25VLLineColor}] plot coordinates {(0,0)};
        Qwen2.5-VL
    };
    \node[font=\scriptsize, text=YOLOWorldLineColor, anchor=west] at (axis cs:3.4, 54) {
        \tikz[baseline=-0.5ex] \path[mark=diamond*, mark size=2pt, mark options={fill=white, line width=0.6pt, draw=YOLOWorldLineColor}] plot coordinates {(0,0)};
        YOLO-World
    };
    \end{axis}
  \end{tikzpicture}

  \vspace{-2.0ex} 

  \begin{tikzpicture}
    \begin{axis}[
        common axis style,
        ylabel={($\mathcal{D}_2$)\\mIoU(\%)},
        ymax=85,
        xmax=5.5, 
        extra description/.code={
            \draw[black, line width=1pt] (axis cs:\commonxmin,0) -- (axis cs:5.5,0);
            \node[font=\scriptsize, color=plot2color, anchor=base] at (axis cs:1, \stepBaselineY) {Step$_1$};
            \node[font=\scriptsize, color=plot2color, anchor=base] at (axis cs:2, \stepBaselineY) {Step$_2$};
            \node[font=\scriptsize, color=plot2color, anchor=base] at (axis cs:3, \stepBaselineY) {Step$_3$};
            \node[font=\scriptsize, color=plot2color, anchor=base] at (axis cs:4, \stepBaselineY) {Step$_4$};
            \node[font=\scriptsize, color=plot2color, anchor=base] at (axis cs:5, \stepBaselineY) {Step$_5$};
        }
    ]
    \addplot[florence2 plot style] coordinates {
        (1, 80.1) (2, 68.3) (3, 55.2) (4, 42.5) (5, 38.5)
    };
    \addplot[groundingdino plot style] coordinates {
        (1, 79.6) (2, 64.8) (3, 52.9) (4, 40.7) (5, 30.5)
    };
    \addplot[qwen25vl plot style] coordinates {
        (1, 78.3) (2, 63.5) (3, 51.9) (4, 44.8) (5, 29.7)
    };
    \addplot[yoloworld plot style] coordinates {
        (1, 78.1) (2, 51.3) (3, 32.1) (4, 21.3) (5, 11.2)
    };
    \end{axis}
  \end{tikzpicture}

  \vspace{-2.0ex}

  \begin{tikzpicture}
    \begin{axis}[
        common axis style,
        ylabel={($\mathcal{D}_3$)\\mIoU(\%)},
        ymax=85,
        xmax=\commonxmax, 
        extra description/.code={
            \draw[black, line width=1pt] (axis cs:\commonxmin,0) -- (axis cs:\commonxmax,0);
            \node[font=\scriptsize, color=plot3color, anchor=base] at (axis cs:1, \stepBaselineY) {Step$_1$};
            \node[font=\scriptsize, color=plot3color, anchor=base] at (axis cs:2, \stepBaselineY) {Step$_2$};
            \node[font=\scriptsize, color=plot3color, anchor=base] at (axis cs:3, \stepBaselineY) {Step$_3$};
            \node[font=\scriptsize, color=plot3color, anchor=base] at (axis cs:4, \stepBaselineY) {Step$_4$};
            \node[font=\scriptsize, color=plot3color, anchor=base] at (axis cs:5, \stepBaselineY) {Step$_5$};
            \node[font=\scriptsize, color=plot3color, anchor=base] at (axis cs:6, \stepBaselineY) {Step$_6$};
        }
    ]
    \addplot[florence2 plot style] coordinates {
        (1, 80.1) (2, 68.3) (3, 55.2) (4, 42.5) (5, 38.5) (6, 33.0)
    };
    \addplot[groundingdino plot style] coordinates {
        (1, 79.6) (2, 64.8) (3, 52.9) (4, 40.7) (5, 30.5) (6, 24.0)
    };
    \addplot[qwen25vl plot style] coordinates {
        (1, 78.3) (2, 63.5) (3, 51.9) (4, 44.8) (5, 29.7) (6, 23.5)
    };
    \addplot[yoloworld plot style] coordinates {
        (1, 78.1) (2, 51.3) (3, 32.1) (4, 21.3) (5, 11.2) (6, 9.0)
    };
    \end{axis}
  \end{tikzpicture}

   \vspace{-1.5ex}
   \caption{Step-by-Step Evaluation of VLMs on VG Task.}
   \label{fig:mean_iou_all_models}
   \vspace{-5ex} 
\end{figure}

%% file: sec/5_conclusion.tex
\vspace{-1em}
\section{Conclusion}
In this work, we define and formalize perception-to-reasoning task complexity to address the absence of a reasoning complexity definition in existing datasets. We develop an adaptive query engine that dynamically generates non-predefined questions with adjustable complexity, enabling dataset creation with controlled question difficulty and task customization. We also provide structured, step-by-step reasoning annotations. In addition, we labeled human-object interactions manually and computing geometric relationships, thereby enhancing the robotic JRDB dataset for integration with the query engine. Our engine and benchmark support fine-grained evaluation of visual reasoning frameworks and dynamic assessment of VLMs across multiple reasoning levels.